\documentclass{IEEEcsmag}

\usepackage[colorlinks,urlcolor=blue,linkcolor=blue,citecolor=blue]{hyperref}
\usepackage{xurl}
\usepackage{upmath}

\jvol{XX}
\jnum{XX}
\paper{8}
\jmonth{May/June}
\jname{IT Professional}
\pubyear{2019}

\begin{document}

\sptitle{Department: Computing and Mathematics}
\editor{Editor: Bill Cassidy, B.Cassidy@mmu.ac.uk and Moi Hoon Yap, M.H.Yap@mmu.ac.uk}

\title{A Cloud-based Deep Learning Framework for Remote Detection of Diabetic Foot Ulcers}

\author{Bill Cassidy}
\affil{Manchester Metropolitan University, UK}

\author{Neil D. Reeves}
\affil{Manchester Metropolitan University, UK}

\author{Joseph M. Pappachan}
\affil{{Lancashire Teaching Hospitals NHS Foundation Trust, Manchester Metropolitan University and The University of Manchester, UK}}

\author{Naseer Ahmad}
\affil{Manchester University NHS Foundation Trust, UK}

\author{Samantha Haycocks}
\affil{Salford Royal NHS Foundation Trust, UK}

\author{David Gillespie}
\affil{Manchester Metropolitan University, UK}

\author{Moi Hoon Yap}
\affil{Manchester Metropolitan University, UK}

\markboth{Department Head}{Paper title}

\begin{abstract}
This research proposes a mobile and cloud-based framework for the automatic detection of diabetic foot ulcers and conducts an investigation of its performance. The system uses a cross-platform mobile framework which enables the deployment of mobile apps to multiple platforms using a single TypeScript code base. A deep convolutional neural network was deployed to a cloud-based platform where the mobile app could send photographs of patient's feet for inference to detect the presence of diabetic foot ulcers. The functionality and usability of the system were tested in two clinical settings: Salford Royal NHS Foundation Trust and Lancashire Teaching Hospitals NHS Foundation Trust. The benefits of the system, such as the potential use of the app by patients to identify and monitor their condition are discussed.
\end{abstract}

\maketitle

\chapterinitial{Diabetes Mellitus} is a chronic metabolic disorder, and a growing world-wide epidemic \cite{kharroubi2015epidemic}. Diabetic Foot Ulcers (DFU) are wounds developed on the feet that represent serious chronic complications resulting from diabetes, and are prone to high levels of recurrence and infection \cite{jia2017infection}. There are numerous potential contributing factors to the development of DFU, with diagnosis, monitoring, and treatment programmes requiring multidisciplinary medical expertise. Feet of diabetic patients are more susceptible to injury and chronic wounds, resulting in skin damage and ultimately development of a DFU \cite{rani2017contour}.

Patients with an active DFU, or at high-risk of developing a DFU require frequent foot checks by healthcare professionals and referral to specialists to prevent additional severe complications \cite{prompers2008delivery}. DFU can result in serious lifestyle repercussions, resulting in immobility, social stigma, social isolation, increased mortality, and significant costs to healthcare systems \cite{jeffcoate2018prevention}, with hospitalisation constituting the most expensive part of treating DFU infections \cite{lipsky2004diagnosis}. More than half of DFUs become infected, with approximately 20\% of moderate or severe DFU infections leading to lower extremity amputation \cite{armstrong2017recurrence}.

The cost of health care in England for DFU and amputation in 2014-2015 is estimated at £1 billion. This constitutes approximately 1\% of the entire National Health Service (NHS) budget \cite{kerr2017diabetic}. The lower bound of DFU and associated amputation cost estimates is higher than the combined NHS expenditure in England on breast, prostate, and lung cancers \cite{kerr2019cost}. In the United States, the direct costs of treating DFU exceed the treatment costs of many common cancers \cite{armstrong2017recurrence}.

Given the significant and growing impact of DFU, mobile health solutions that target this condition could assist in improving patient quality of life. Up to 80\% of DFU are thought to be preventable through early detection \cite{machin2017thermal}. Promotion of patient self care and continuous monitoring for those most at risk, increasing rates of early intervention to reduce the severity and impact of DFU, could provide significant cost savings for healthcare systems. Self-management programmes have been associated with improved health outcomes, with mobile technologies identified as an important factor in helping to deliver self-management interventions that are adaptable, low cost and easily accessible \cite{whitehead2016apps}.  

 
Due to the continued significant increase in reported global cases of diabetes and DFU, research in this area has also seen notable growth. As a result, the use of deep learning algorithms for automated analysis of DFU have become more prominent, particularly from our group over the last few years \cite{goyal2017dfunet, goyal2017fully, goyal2018robust, cassidy2020dfuc}. Goyal et al. have created and validated deep convolutional neural networks (CNNs) capable of DFU classification \cite{goyal2017dfunet}, semantic segmentation \cite{goyal2017fully} and localisation \cite{goyal2018robust}. These models have been shown to have high levels of sensitivity, specificity and mean average precision (mAP) in experimental settings.

This paper proposes a cloud-based deep learning framework for remote detection of diabetic foot ulcers. To address the issues above, our framework includes:

\begin{itemize}
    \item A cross-platform mobile app used for capturing photographs of DFU (a non-contact solution) capable of sending diagnosis requests to a cloud service
    \item A cloud-platform that mobile clients can connect to capable of inference using one or more CNNs to provide a diagnosis
\end{itemize}

To assess the usability and reliability of such a system, we completed a proof-of-concept clinical evaluation using mobile and cloud technologies at two UK sites: Salford Royal NHS Foundation Trust and Lancashire Teaching Hospitals NHS Foundation Trust. Prior to starting the proof-of-concept clinical evaluation, we obtained ethical approval from Salford Royal NHS Foundation Trust (REF: S19HRANA37) and Lancashire Teaching Hospitals NHS Foundation Trust (REF: SE-281).

\section{WHY CLOUD?}
The unprecedented growth of the global smartphone market over the last decade has been mirrored by the more recent emergence and rapid expansion of enterprise Cloud Computing Platforms (CCP). CCP provide on-demand computing, storage and software accessible over the internet, allowing for the remote offloading of process-intensive tasks. This approach to server technology is an increasingly common long-term strategy for replacing the traditional manually maintained client-server hardware set-up \cite{labati2020cloud}.


A clear advantage of CCPs are that they allow users of mobile devices to gain access to significant processing power, well beyond the means of any existing mobile device. This approach allows for patients to use even very dated mobile hardware to access the latest advances in automated medical image analysis. This essentially means that continual advances in this field are not tied to the computing capability of mobile devices, as such devices are simply consuming services from CCPs. Additionally, scalability becomes easier to manage, given the virtualised nature of cloud services. There is a growing trend in the use of ensemble CNNs in medical image analysis, whereby multiple CNNs are used to form a final prediction. Distributing mobile apps that use multiple models is not practical or possible given the limited permissible size of apps when distributed via online app stores. There is also the issue of intellectual property protection. Android apps are particularly easy to reverse engineer, so having the CNNs run on the server instead of the user's mobile device means that trained models are never publicly exposed.


\section{SYSTEM ARCHITECTURE}
\begin{figure*}
	\centering
	\includegraphics[scale=0.56]{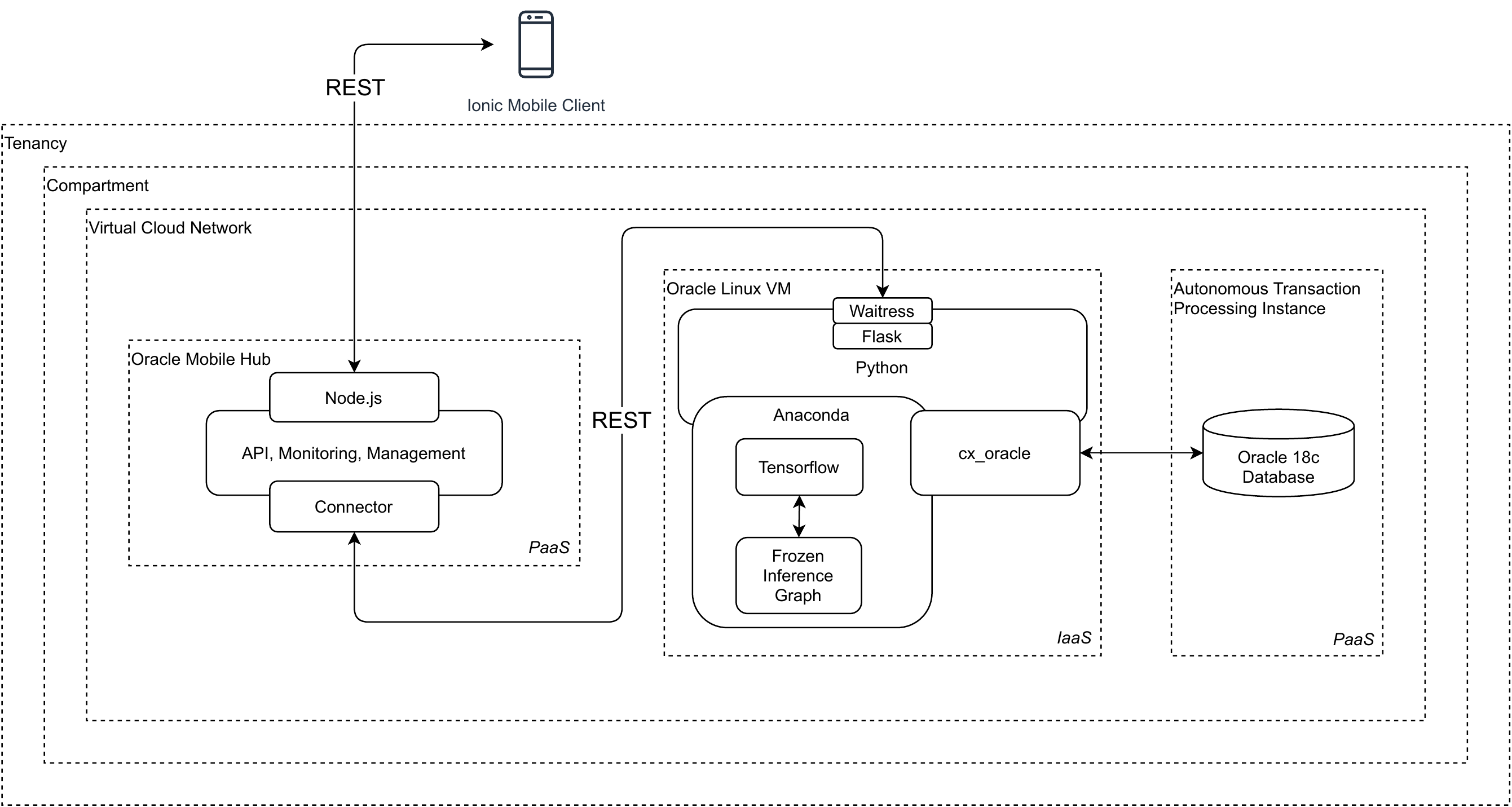}
	\caption{An overview of the physical architecture showing the major structural components involved in the implementation.}
	\label{fig:PhysArchitecture}
\end{figure*}

The two major components created for the evaluation were (1) a cross-platform mobile app, and (2) a cloud-based deep learning framework that performed inference on foot photographs sent from mobile clients. A cross-platform framework was chosen for the development of the mobile client since the ultimate goal of this research is to provide patients with a means of remotely monitoring and diagnosing DFU using their own smart-phones, which primarily comprise of Android or iOS devices. An overview of the system physical architecture is shown in Fig. \ref{fig:PhysArchitecture}. The following sections describe how these components were utilised in the creation of our proposed framework.


\subsection{Mobile App}
\begin{figure*}
	\centering
	\includegraphics[scale=0.37]{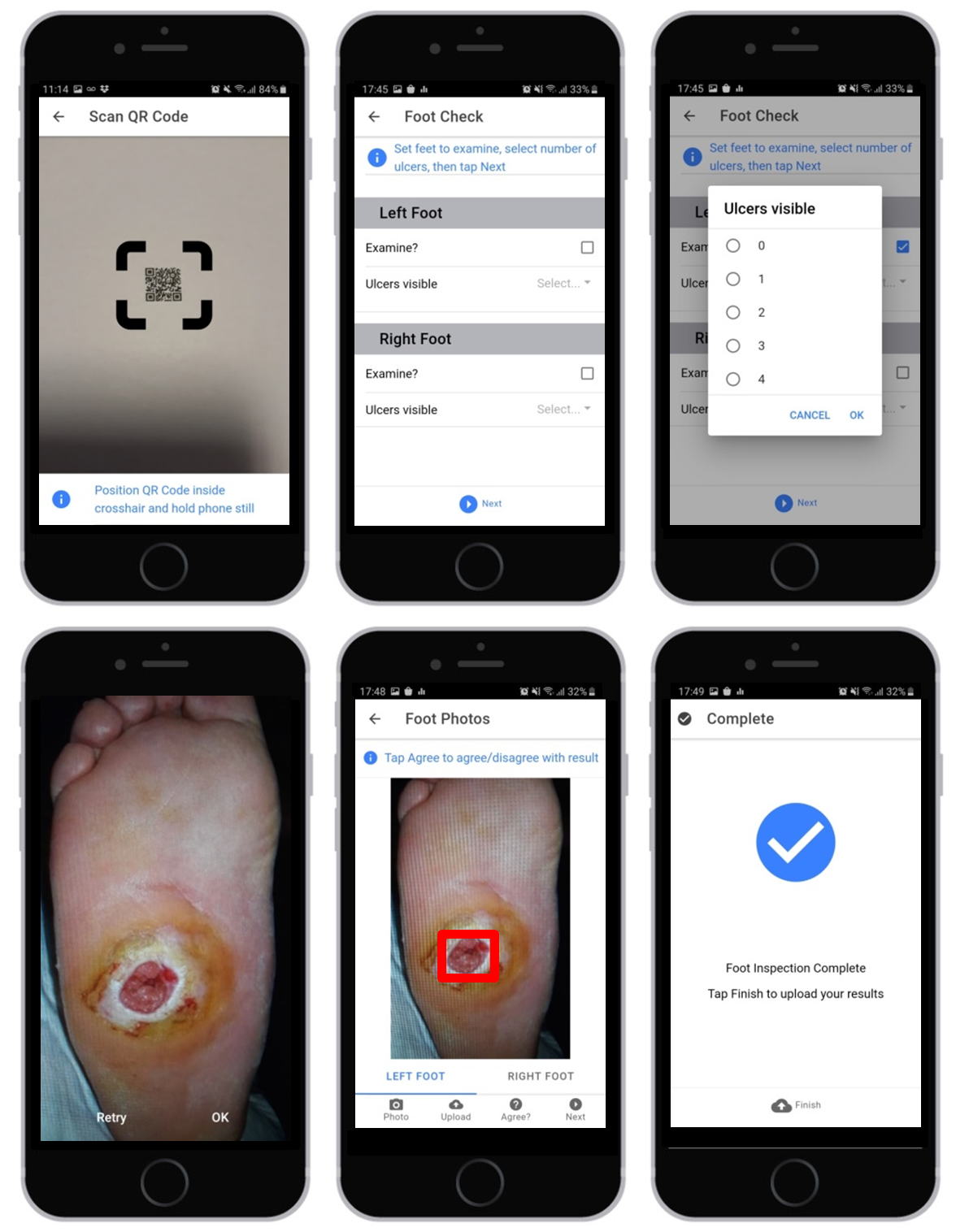}
	\caption{UI screenshots from the proposed cross-platform mobile client. From left to right, (top-left) patient QR code is scanned, (top-middle) clinician enters details of each foot being examined, (top-right) clinician enters number of visible ulcers, (bottom-left) photo acquisition of foot, (bottom-middle) inference result returned by cloud service, (bottom-right) examination complete.}
	\label{fig:MobileAppScreens}
\end{figure*}

Cross-platform development can help to reduce the time and costs associated with developing apps for multiple mobile platforms. The mobile app developed for our evaluation was created using Ionic, a cross-platform framework based on the earlier Cordova framework. Screens within Ionic apps are rendered onto a standard WebView, in the same way that web pages are rendered in web browsers. There are also native elements within the framework however, including the ability to interface with the device's hardware features, such as sensors and cameras. Fig. \ref{fig:MobileAppScreens} shows the main data capture screens within the mobile app.

The primary objective of our initial proof of concept evaluation was to determine the usability and reliability of our cross-platform mobile client and cloud-based framework in real-world settings. Ease of use was a primary motivating factor behind the design of the mobile app. Screens within the app display context-sensitive information in the form of an information bar at the top of each screen that was used to guide the user through the process of acquiring and uploading foot photographs. The UI and validation were designed so that it was not possible for the user to take the wrong action. Examples of this include: 

\begin{itemize}
	\item It was not possible to retake a photograph for the current foot if one had already been taken and uploaded.
	\item The user could not upload a photograph for any foot more than once.
	\item It was not possible to change left foot ``checked" tickbox if the left foot photo had been uploaded.
	\item It was not possible to change right foot ``checked" tickbox if the right foot photo had been uploaded.
\end{itemize}

Ionic utilises a Model View Controller (MVC) architecture, implemented using Angualr.js, which separates data, presentation of data and business logic. App data, including application state, is stored in a local SQLite database.

\subsection{Oracle Mobile Cloud Service Software Development Kit (SDK)}
Oracle provides a SDK for several mobile development platforms, including Ionic, which enables mobile clients to interface with Oracle Mobile Hub (OMH). The Oracle Mobile Cloud Service SDK is a HyperText Transfer Protocol Secure client layer, through which requests can be made to OMH and associated services using JavaScript Object Notation (JSON) via REpresentational State Transfer (REST) to transfer data between clients and the cloud service.

\subsection{Cloud Platform}
The cloud platform services developed for our proof-of-concept clinical evaluation were implemented using Oracle Cloud Infrastructure (OCI). OCI is an online enterprise scale cloud service offering Infrastructure as a Service (IaaS), Platform as a Service (PaaS) and Software as a Service (SaaS). A breakdown of these service models are described in the following sub-sections.

\subsubsection{Platform as a Service}
Oracle Mobile Hub (OMH) and the Autonomous Transaction Processing Instance (ATPI) represent the PaaS elements used in the evaluation. OMH provides a gateway for mobile clients to access other internal cloud services, and includes features such as identity management, analytics, and Application Programming Interface (API) management.


The ATPI hosts the Oracle 18c database, which is used for storing all data relating to the evaluation, including foot details entered by clinicians, photographs taken during patient appointments, inference results returned from the deep learning model and clinician confirmation of agreement with inference results. ATPI offers multiple deployment options that automatically configures the database depending on its targeted use case. For our evaluation, the ATPI was configured to use the Autonomous Transaction Processing workload type, which optimises the database with a bias towards processing high volumes of random data access.


\subsubsection{Infrastructure as a Service}
Oracle Compute represents the IaaS component of the project, consisting of a Virtual Machine (VM). Virtualisation software allows multiple guest systems to run on a single physical platform, where isolated environments can be created by multiplexing host's computing cycles and virtualising hardware resources. The VM hosts the core of the business logic, together with the frozen inference graph used for inference. For our proof-of-concept clinical evaluation, the operating system used was Ubuntu 16.04.6 LTS (xenial) with Nvidia GPU Cloud Machine Image shape, which defines the hardware configurations that are available to the VM instance. Hardware available on this shape included an Intel Xeon Gold 5120 2.20GHz CPU, and an Nvidia Tesla P100 SXM2 16GB GPU. We created two python programs to run on the VM, which were responsible for processing network, database, and image inference operations.

The first of the two Python programs, ServerPy, handled incoming requests from mobile clients via OMH over REST. All incoming requests are handled by Flask - a Python web framework that allows for the routing of incoming REST requests to Python classes. Requests are made to either add new data to the database, update existing data, or to retrieve data from the database. Adding data to the database includes adding new patient foot data, patient foot photographs, and clinician agreement confirmation with inference results. Sending data from the database to requesting clients takes the form of server status codes, app version checks to ensure the user is using the correct version of the mobile app, and the results of completed inference requests. Additionally, when new photographs are received by ServerPy, the details are added to a jobs table in the ATPI database. The second Python program, AnnotatePy, was responsible for periodically reading the jobs table and retrieving the oldest incomplete job. The job is then processed, using TensorFlow for inference. Once inference is completed, the results are added to the database, and the job is marked as complete. This process operates as a queue, using a first in first out principle.



\subsection{Deep Learning Framework}
The DFU localisation model trained by Goyal et al. \cite{goyal2018robust} was selected for use during our proof-of-concept clinical evaluation. This single classifier model showed the highest mAP (91.8) in a comparison of supervised deep learning models trained and evaluated with DFU. The model was trained using 1,775 DFU images, with ground truth labelling provided by clinical diabetic foot experts at Lancashire Teaching Hospitals NHS Foundation Trust. It implements Faster R-CNN as the object localisation network to process feature extraction, with Inception-ResNetV2 used to classify the extracted feature maps. This model was trained using two-tier (partial and full) transfer learning using the MS COCO dataset, implements three distinct steps to perform localisation:

\begin{enumerate}
	\item Feature extraction using Inception V2 which serves as input for later stages (proposals and classifier).
	\item Generation of proposals and refinement.
	\item Region of interest classifier and bounding box regressor to fine-tune bounding box accuracy.
\end{enumerate}

The model was trained using a heterogeneous dataset consisting of non-standardised images of DFU. Aspects such as orientation, distance from foot, capture device type, resolution, focal length, exposure time, ISO speed ratings, variances in the amount of the foot visible in the image, and lighting conditions resulted in a high level of variability in image characteristics. It could be argued that a non-standardised dataset is more desirable in the training of such a model, since this would increase the viability of its use in real world settings where a system would need to be able to take into account numerous uncontrolled environment variables.

\begin{table*}
\caption{Summary of Results for Individual Question Responses, reported in mean $\pm$ SD (standard deviation).}
\label{tab:questions}
\scalebox{1.0}{
\begin{tabular}{|l|l|l|}
\hline
    & Question & mean$\pm$SD \\ \hline
 Q1 & The app was easy to use & 6.50$\pm$0.55\\  \hline
 Q2 & It was easy for me to learn to use the app & 6.83$\pm$0.41\\  \hline
 Q3 & The navigation was consistent when moving between screens & 5.83$\pm$0.98\\  \hline
 Q4 & The interface of the app allowed me to use all the functions offered by the app & 6.00$\pm$0.89\\  \hline
 Q5 & Whenever I made a mistake using the app, I could recover easily and quickly & 4.75$\pm$2.22\\  \hline
 Q6 & I like the interface of the app & 5.83$\pm$0.98\\  \hline
 Q7 & The information in the app was well organised, so I could easily find the information I needed & 6.00$\pm$1.27\\  \hline
 Q8 & The app adequately acknowledged and provided information to let me know the progress of my action & 5.00$\pm$2.35\\  \hline
 Q9 & Overall, I am satisfied with this app & 6.00$\pm$1.27\\  \hline
 Q10 & This app has all the functions and capabilities I expected it to have & 4.80$\pm$2.28\\  \hline
\end{tabular}
}
\end{table*}

\section{FINDINGS}
Usability is a key factor in the adoption of mobile health apps, especially in cases where users are not within the typical age range of mobile users \cite{zapata2015mhealth}. Therefore, at the end of our six-month proof-of-concept evaluation, users of the system, clinicians, were asked to complete a usability questionnaire. The University of Pittsburgh Usability Questionnaire (Standalone Mobile Health App for Health Care Providers template) \cite{pitt2019usability} was used to obtain the qualitative measures for our evaluation. Zhou et al. validated the PITT Usability Questionnaire, and it was shown to have high internal consistency reliability \cite{zhou2019usability}. Questions that were not relevant to the app in its current state, such as questions 9, 15 and 18 were excluded as they were considered not to be relevant to the use of the app in its current prototype stage. We also added an additional free-text section at the end of the questionnaire asking clinicians to provide any other details of their experience using the app, and any recommendations for improvement. Six participating clinicians completed the questionnaire, with questions scored between 1 and 7; 1 being disagree and 7 being agree. They were also able to select a not applicable option if they believed that the statement did not relate to themselves. 

\subsection{Quantitative Analysis}
Table \ref{tab:questions} shows a summary of mean and standard deviation for the ratings of each question. The questionnaire results indicate that all participating clinicians report high levels of satisfaction when using the app, with most of the mean scores being above 5, of a maximum score of 7. Questions 1 (m = 6.50; sd = 0.55) and 2 (m = 6.83; sd = 0.41), which relate to ease of use, provide the highest scores, which we regard as a good indicator that the app would be easy for patients to use in home settings, and meets one of the main criteria when taking into account the design of the app. The lowest scoring questions were Question 5 (m = 4.75; sd = 2.22) and Question 10 (m = 4.80; sd = 2.28), which related to how the app responds to user mistakes and expected app functionality respectively. This would indicate that the app design might benefit from further adjustments to enable users to more easily correct their mistakes. However, these issues would be negated in a patient-focused app since it would not contain any of the data entry elements currently present in the clinician-focused app prototype.

\subsection{Qualitative Analysis}
The free text responses provided by participating clinicians showed varying results that were not obvious to gauge from the answers to the Likert scale questions. Most participating clinicians were in agreement that the app was easy to use and functioned as expected. However, some clinicians experienced connectivity issues with the app due to the restrictive nature of free hospital WiFi, which resulted in occasionally slow upload of foot photographs. Such restrictions may mean that connected devices are automatically disconnected after a period of inactivity, with the only way to reconnect being via the device web browser, a process that has to be completed manually by the user.

Clinicians also agreed that when the app detected an ulcer, the ulcer localisation results were generally highly accurate. Other responses noted the number of false positive detections, with clinicians indicating that such detections would occur around callouses or extravasation areas around the wound. However, one response noted that extravasations detected as ulcers would at least direct patients to the clinic for assessment. One response noted that the app would be less useful for clinicians in its current state as they already knew how to recognise the presence of an ulcer. However, other responses disagreed with this statement, highlighting the importance of regular photographic capture of DFUs for screening and remote serial analysis. A device that allows patients to self-screen at home could encourage diabetic patients to check their feet more regularly, and would enable clinicians to check patients' feet without the need for hospital visits. It was also noted that older patients might have difficulty using the app without assistance. This could be addressed with the help of a partner, family member or carer. Another solution would be to use of a selfie-stick attachment for the mobile device.






\section{Recommendations and Future Work}
The two Python programs that run on the VM, ServerPy and AnnotatePy, might be better deployed as Linux services, so that they are able to automatically recover from system crashes and other system malfunctions. For the mobile app, in the case of a larger scale patient-focused evaluation, it might be preferable to implement push notifications so that users do not have to keep the app open while waiting for the cloud service to return inference results. A form of data protection might also be employed to ensure that all data and photographs are encrypted on the device. Given that an internet connection is required for the app to function, encryption of all local data could be achieved by requesting encryption keys from the cloud service as and when needed, but without ever actually storing the keys on the device itself.


A larger evaluation might require a more robust server-side solution, where multiple GPUs could be employed to complete a larger volume of jobs simultaneously. This could be achieved by sending each job to a single GPU, or distributing the workload of multiple jobs to multiple GPUs simultaneously. Further investigation is required to determine the most performant solution in this scenario.

Our framework has been designed to encourage frequent patient self-monitoring, supporting early detection of DFU that will lead to earlier signposting to treatment and therefore improved ulcer healing. Early intervention is an important factor in improving healing rates. Tools and education programmes to give patients the knowledge and motivation to manage their condition are essential in the goal of reducing the negative effects of diabetes and DFU \cite{boodoo2017mhealth}. 

Many people diagnosed with DFU are older adults, therefore it will be important to ensure that any future apps created for use with our framework are extremely simple and easy to use. They should require the absolute minimum input from the user, and results should be presented in a form that are easy to understand. Usability will be the primary defining objective for a patient-focused version of the app. Minimal complexity will ensure the greatest adoption and impact of the system.

In the current system implementation, foot photographs are uploaded and stored in an Oracle 18c database as Binary Large Objects, together with all the other data captured during the evaluation. In the context of an initial evaluation, this approach was deemed appropriate, given that the average photograph file size was 60KB. However, in the context of a larger study, it may be beneficial to separate photographs and database data. Excessive file storage within the database may impact its performance over time. Photographs could instead be saved to the VM file system. However, this may require careful design to ensure that photographs are archived and easily retrievable, which would likely involve managing directories. A more desirable alternative might be to utilise a cloud storage layer, where photographs are given unique IDs and are sent to the storage layer which removes the need to manage file system elements. Photographs can be easily retrieved from the storage layer using an API and the photograph UID. Further ahead, the amount of data collected by our system could increase exponentially, especially if used internationally. Such datasets would classify as Big Data, whereby a NoSQL database may prove to be a superior fit for such large volumes of data processing.

During the analysis phase of the project, we explored the possibility of using a serverless solution, whereby the setup of a VM to host the Python applications could be bypassed. Instead, packages would be uploaded to a server space where application methods can be triggered by events received via REST requests. However, this approach to cloud computing is still in the early stages, with most providers not exposing access to GPU resources using this method.

Following the positive results in user acceptance from our proof-of-concept evaluation, we plan for a larger scale study to be undertaken. This follow-up study will be patient-focused, where the app will be simplified and distributed to a larger number of users. In this study, the app will be used by patients, their friends, family, or carers, instead of clinicians. This next stage will provide confirmation of whether the app and associated technologies are suitable for large-scale real-world use. The technologies developed will form the basis of a platform to support future research into areas such as:

\begin{itemize}
	\item Automated early detection of DFU, including the detection of signs of pre-ulceration.
	\item Automated classification and segmentation of DFU types: (1) DFU with no infection and no ischemia, (2) DFU with infection and no ischemia, (3) DFU with ischemia and no infection,  (4) DFU with infection and ischemia.
	\item Automated segmentation of DFU tissue types determined by colour and texture features, including necrotic, epithelial, granulation and slough.
	\item Automated non-contact methods of monitoring DFU healing status over time.
	\item Automated non-contact methods of monitoring the periwound (surrounding tissue of a wound) as a potential indicator of wound healing.
	\item Automated non-contact methods of monitoring DFU edemia - a condition characterised by the swelling of all or parts of the foot.
	\item Automated non-contact methods used for DFU pathophysiology.
\end{itemize}

\section{CONCLUSION}
In this work, we developed a cross-platform mobile app and a cloud-based deep learning framework for the automatic detection of DFU. The system was assessed for usability via qualitative methods, which showed that the system scored highly for system usability when used by clinicians in clinical settings. This work will provide the basis for a more extensive patient-focused evaluation of the system to determine its effectiveness when used by patients and their carers. The dataset obtained over the six-month evaluation period will be used to retrain the existing deep learning model to improve its effectiveness in detecting DFU at various stages of development. The longitudinal data will be used to form the basis of a refined model that will be used to detect the early signs of DFU.

To the best of our knowledge, the framework created for this research is the first of its kind, where DFU can be automatically detected and localised by a fully integrated framework of state-of-the-art technologies with an easy to use app, producing high confidence scores, where inference is performed in the cloud. This could lead to the eventual expansion of our system for use as a tool, not just for patients to self-monitor, but also as a diagnosis tool for medical experts. Our framework can now be used as a platform for further research, including early detection of DFU, and monitoring of DFU healing status over time. Further ahead, our platform could be expanded into other areas of research and automatic medical wound analysis, including other pathologies, on any part of the human body. 

\section{ACKNOWLEDGMENT}

The authors would like to thank Oracle Research for providing the IaaS and PaaS technologies that enabled our clinical evaluation to take place. Gratitude is also extended to Salford Royal NHS Foundation Trust, Lancashire Teaching Hospitals NHS Foundation Trust and Manchester University NHS Foundation Trust for their extensive support during the usability study. This project is funded by The Manchester Metropolitan Strategic Operation Fund and Oracle Innovator Accelerator Programme.


\begin{thebibliography}{1}

\bibitem{kharroubi2015epidemic}
A.~T. Kharroubi and H.~M. Darwish, ``Diabetes mellitus: The epidemic of the century,''
\emph{World Journal of Diabetes}, vol.~6, 2015.

\bibitem{jia2017infection}
L.~Jia, C.~N. Parker, T.~J. Parker, E.~M. Kinnear, P.~H. Derhy, A.~M. Alvarado, F.~Huygens, and P.~A. Lazzarini, ``Incidence and risk factors for developing infection in patients presenting with uninfected diabetic foot ulcers,''
\emph{PLoS One}, vol.~12, no.~5, 2017.

\bibitem{rani2017contour}
P.~Rani, B.~Aliahmad, and D.~K. Kumar, ``A novel approach for quantification of contour irregularities of diabetic foot ulcers and its association with ischemic heart disease,'' in \emph{39th Annual International Conference of the IEEE Engineering in Medicine and Biology Society (EMBC)}, 2017, pp.1437--1440.

\bibitem{prompers2008delivery}
L.~Prompers, M.~Huijberts, J.~Apelqvist, E.~Jude, A.~Piaggesi, K.~Bakker,
M.~Edmonds, P.~Holstein, A.~Jirkovska, D.~Mauricio \emph{et~al.}, ``Delivery of care to diabetic patients with foot ulcers in daily practice: results of the eurodiale study, a prospective cohort study,'' \emph{Diabetic medicine}, vol.~25, no.~6, pp. 700--707, 2008.

\bibitem{jeffcoate2018prevention}
W.~J. Jeffcoate, L.~Vileikyte, E.~J. Boyko, D.~G. Armstrong, and B.~A. JM,
``Current challenges and opportunities in the prevention and management of diabetic foot ulcers,'' \emph{Diabetes Care}, vol.~41, pp. 645--652, 2018.

\bibitem{lipsky2004diagnosis}
B.~A. Lipsky, A.~R. Berendt, H.~G. Deery, J.~M. Embil, W.~S. Joseph, A.~W.
Karchmer, J.~L. LeFrock, D.~P. Lew, J.~T. Mader, C.~Norden \emph{et~al.},
``Diagnosis and treatment of diabetic foot infections,'' \emph{Clinical 
Infectious Diseases}, vol.~39, no.~7, pp. 885--910, 2004.
	
\bibitem{armstrong2017recurrence}
D.~G. Armstrong, A.~J. Boulton, and S.~A. Bus, ``Diabetic foot ulcers and their recurrence,'' \emph{The New England Journal of Medicine}, vol. 376, pp.2367--2375, 2017.

\bibitem{kerr2017diabetic}
M.~Kerr, ``Diabetic foot care in england: an economic study,'' 2017, last
access: 08/11/19. [Online]. Available: \url{https://www.evidence.nhs.uk/document?id=1915227&returnUrl=Search%3Fq%3DAmputation%2Bcost&q=Amputation+cost}
	
\bibitem{kerr2019cost}
M.~Kerr, E.~Barron, P.~Chadwick, T.~Evans, W.~Kong, G.~Rayman, M.~Sutton-Smith, G.~Todd, B.~Young, and J.~WJ, ``The cost of diabetic foot ulcers and amputations to the national health service in england,'' \emph{Diabetic Medicine}, vol.~36, pp. 995--1002, 2019.	
	
\bibitem{machin2017thermal}
G.~Machin, A.~Whittam, S.~Ainarkar, J.~Allen, J.~Bevans, M.~Edmonds, B.~Kluwe, A.~Macdonald, N.~Petrova, P.~Plassmann, F.~Ring, L.~Rogers, and R.~Simpson, ``A medical thermal imaging device for the prevention of diabetic foot ulceration,'' \emph{Physiol Meas.}, vol.~38, no.~3, pp. 420--430, 2017.
	
\bibitem{whitehead2016apps}
L.~Whitehead and P.~Seaton, ``The effectiveness of self-management mobile phone and tablet apps in long-term condition management: A systematic review,''
\emph{Journal of Medical Internet Research}, vol.~18, no.~5, May 2016.
	
\bibitem{goyal2017dfunet}
M.~Goyal, N.~D. Reeves, A.~K.Davison, S.~Rajbhandari, J.~Spragg, and M.~H. Yap, ``DFUNet: Convolutional Neural Networks for Diabetic Foot Ulcer Classification'' \emph{IEEE Transactions on Emerging Topics in Computational Intelligence}, vol.~4, pp. 728--739, 2018.
			
\bibitem{goyal2017fully}
M.~Goyal, M.~H. Yap, N.~D. Reeves, S.~Rajbhandari, and J.~Spragg, ``Fully
convolutional networks for diabetic foot ulcer segmentation,'' in \emph{2017 IEEE International Conference on Systems, Man, and Cybernetics (SMC)}, Oct 2017, pp. 618--623.
			
\bibitem{goyal2018robust}
M.~Goyal, N.~Reeves, S.~Rajbhandari, and M.~H. Yap, ``Robust methods for
real-time diabetic foot ulcer detection and localization on mobile devices,''
\emph{IEEE journal of biomedical and health informatics}, 2018.

\bibitem{cassidy2020dfuc}
B.~Cassidy, N.~D. Reeves, P.~Joseph, D.~Gillespie, C.~O'Shea, S.~Rajbhandari, A.~G. Maiya, E.~Frank, A.~Boulton, D.~Armstrong, B.~Najafi, J.~Wu and M.~H. Yap, 
``DFUC2020: Analysis Towards Diabetic Foot Ulcer Detection, ''
\emph{arXiv preprint arXiv:2004.11853}, 2020.

\bibitem{labati2020cloud}
R.~D. Labati, A.~Genovese, V.~Piuri, F.~Scotti, and S.~Vishwakarma,
\emph{Recent Advances in Intelligent Engineering}.\hskip 1em plus 0.5em minus 0.4em\relax Springer, Cham, 2020.

\bibitem{zapata2015mhealth}
B.~Zapata, J.~Fernández-Alemán, A.~Idri, and A.~Toval, ``Empirical studies on usability of mhealth apps: a systematic literature review,'' \emph{J Med Syst.}, vol.~39, no.~2, 2015.
			
\bibitem{pitt2019usability}
``{PITT Usability Questionnaire},'' 2019, last access: 11/11/19. [Online].
Available: \url{https://ux.hari.pitt.edu/v2/portal/#/}

\bibitem{zhou2019usability}
L.~Zhou, J.~Bao, M.~A. Setiawan, A.~Saptono, and B.~Parmanto, ``The mhealth app usability questionnaire (mauq): Development and validation study,'' \emph{JMIR Mhealth Uhealth}, vol.~7, no.~4, 2019.
			
\bibitem{boodoo2017mhealth}
C.~Boodoo, J.~Perry, and P.~e.~a. Hunter, ``Views of patients on using mhealth to monitor and prevent diabetic foot ulcers: Qualitative study,'' \emph{JMIR Diabetes}, vol.~2, no.~2, 2017.

\end{thebibliography}
\end{document}